\documentclass{article}
 \usepackage[a4paper, total={7in, 8in}]{geometry}
\usepackage[utf8]{inputenc} 
\usepackage[T1]{fontenc}    
\usepackage{hyperref}       
\usepackage{url}            
\usepackage{booktabs}       
\usepackage{amsfonts}       
\usepackage{nicefrac}       
\usepackage{microtype}      
\usepackage{lipsum}
\usepackage{bm}
\usepackage{booktabs}
\usepackage{pgfplots}
\usepackage{pgfplotstable}
\RequirePackage[authoryear]{natbib}
\usepackage{subfig}
\usepackage{graphicx}
\usepackage{authblk}

\RequirePackage{amsthm,amsmath}

\title{Sedentary Behavior Estimation with Hip-worn Accelerometer Data: Segmentation, Classification and Thresholding}

\author[1]{Yiren Wang$^1$, Fatima Tuz-Zahra$^1$, Rong Zablocki$^1$, Chongzhi Di$^2$, Marta M. Jankowska$^3$, John Bellettiere$^1$, Jordan A. Carlson$^4$,Andrea Z. LaCroix
 $^1$, Sheri J. Hartman$^1$, Dori E. Rosenberg$^5$, Jingjing Zou$^{1*}$, Loki Natarajan$^{1*}$}
\date{
 $^1$ University of California San Diego\\%
    $^2$ Fred Hutchinson Cancer Research Center\\%
    $^3$ Beckman Research Institute, City of Hope\\
    $^4$ Center for Children's Healthy Lifestyles and Nutrition, Children's Mercy\\
    $^5$ Kaiser Permanente Washington Health Research Institute\\
    $^*$ joint senior authors
}

\begin{document}
	\maketitle
	\begin{abstract}
		
		Cohort studies are increasingly using accelerometers for physical activity and sedentary behavior estimation. These devices tend to be less error-prone than self-report, can capture activity throughout the day, and are economical. 
		However, previous methods for estimating sedentary behavior based on hip-worn data 
		are often invalid or suboptimal under free-living situations and subject-to-subject variation. 
		In this paper, we propose a local Markov switching model 
		that takes this situation into account, and introduce a general procedure for posture classification and sedentary behavior analysis that  
		fits the model naturally. 
		Our method features changepoint detection methods in time series and also a two stage classification step that labels data into 3 classes (sitting, standing, stepping). Through a rigorous training-testing paradigm, we showed that  our approach achieves $\ge 80\%$ accuracy. 
		In addition, our method is robust and easy to interpret.
	\end{abstract}

	\section{Introduction}
	Sedentary behavior (SB) is associated with many health outcomes including cardiovascular disease, obesity, diabetes, clinical depression, and certain types of cancer (\cite{SB01, SB02, SB03,SB04,SB05,SB06,SBHO,SBSHO,pmid19346988, SBECH}). In adults, during awake periods, time spent sitting is the primary contributor to sedentary time. Thus, accurate measurement of sitting versus non-sitting intervals is critical for  assessing associations between sedentary patterns and health, as well as for designing effective interventions to reduce SB.   
	
	Wearable sensors, such as accelerometers are increasingly used to track physical avtivity (PA) and SB \cite{EASB}. 
	These devices are generally economical, scale well to large cohort studies and are not subject to the recall biases inherent in self-report questionnaires. 
	A commonly used approach to characterize PA and SB is to cut-point based metrics of accelerometer-measured counts per minute (CPM). 
	Thresholds for CPM are calibrated to categorize periods of time into sedentary time, light physical activity, and moderate to vigorous physical activity (MVPA) \cite{pmid18303006}. 
	However, this method was found to underestimate total sitting time compared to the actual behavior \cite{KERR2013290, Bellettiere2021}.

	Machine learning methods have been adopted to
	address the limitations of cut-point methods in characterizing PA and SB.
	A random forest algorithm \cite{pmid29443824}, was implemented for the classification of 5-second fixed-length intervals into five different PA/SB categories. \cite{JMPB} proposed a "SedUp" approach to use logistic regression classifiers for wrist-worn activity tracker data. \cite{Nakandala2020} developed a deep learning convolution neural network to classify sitting vs non-sitting periods and compared the prediction accuracy with traditional ML methods. Machine learning methods, with varying degrees of success, were able to accurately classify postures/behaviors. However, they were "black-boxes" approaches focusing primarily on accuracy rather than interpretability. 
	In applications of clinical research and public health, however, interpretability is highly desired for researchers and clinicians to have in-depth understanding of PA and SB behaviors and to make guidelines to promote healthy PA habits.

    In this paper, we propose an interpretable statistical approach: local Markov switching model. We propose to model different kinds of postures (sit, stand, and step) as governed by hidden states and changes in postures as switching among the hidden states.
    We provide here an overview of the key steps in our method. Details on each step are expanded in subsequent sections.

	\begin{itemize}
		\item First, we separate raw tri-axiel accelerometer data into contiguous segments where within each segment the hidden state is believed to be a constant (e.g., a contiguous sitting segment).
		\item Second, we predict the states for each individual segment using a two-step procedure consisting of a highly customized random forest model and multiple hypothesis testing.
		\item Lastly, we correct for spurious predictions obtained in the second step. 
	\end{itemize}

	
	Our proposed method is motivated by and applied to hip-worn triaxial accelerometer data from a cohort of breast cancer survivors \cite{pmid29443824}.
	Numerical results show that our proposed approach has higher classification accuracy 
	comparing to results 
	with the machine learning approach in \cite{pmid29443824};  it is also capable of delivering accurate results for key quantities that describe sedentary behavior, such as total sedentary time and sedentary bout length distribution. 
	
	Our paper is organized as the following. 
	Section 2 describes the data motivated this study and the pre-processing procedure. 
	Section 3 describes in detail the proposal model.
	Section 4 describes the main classification algorithm based on the model. 
	Section 5 provides experimental results of our proposed method and compare results using different methods. 
	Section 6 provides further discussion on the benefits and potential caveats of our method with possible solutions. Conclusions of this paper can be found in Section 7.

	\section{Data description}
	\subsection{Participants and data}
	The study sample involves 28 overweight women breast cancer survivors, who participated in a pilot study examining associations between SB and breast cancer-risk biomarkers; study details have been published \cite{Hartman2018},  \cite{Marinac2018}. Briefly, to be eligible for the study, women had to have been diagnosed with stage I-III breast cancer within 5 years of study enrollment, have completed primary treatment, and had to speak English fluently. At study entry, participants were a mean age of 62 (SD = 7.8) years, on average 2.6 years from their cancer diagnosis, were primarily White (93\%), and over half had completed college. 
	
	Each participant was equipped with a hip-worn Actigraph GT3X+ 
	device and a thigh-worn activPAL device (PAL Technologies, Glasgow, Scotland), with their activities recorded by both devices for seven consecutive days under free-living situation. For convenience and ease of use, the Actigraph GT3X+ device was placed in a pouch that could be worn around the hip. 
	The output files of Actigraph GT3X+ consist of continuous triaxial acceleration measurements measured at granularity of 30 HZ. We use $x,y,z$ to sequentially label the axes in the output files.

	The activPAL device was attached to a fixed location at the front center of the right thigh and therefore can directly estimate limb position, which makes it capable of distinguishing different postures including sitting, standing, and stepping.
	The output activPAL files consist of consecutive "events" that describe  physical postures and lengths of duration, with a granularity of 0.1 second. 
	
	The output files from the two devices were  aligned using the POSIX time information of both devices. 
	Event labels from activPAL (0-sitting, 1-standing, 2-stepping) were used as the ground truth for posture/behavior because in general  thigh-worn devices, although not perfect, are able to accurately measure postures and postural transitions (\cite{doi:10.1080/1091367X.2015.1054390}).  
	Our goal is to develop posture classification algorithms for Actigraph GT3X+ data that have high accuracy comparing to the ActivePal ground truth. Note that upon visual inspection of the time-aligned thigh and hip-worn device recordings, 2 participants are excluded from our analysis due to inconsistencies.

	\subsection{Data Pre-processing}
	
\begin{figure}[h]
		\centering
		\includegraphics[scale = 0.3]{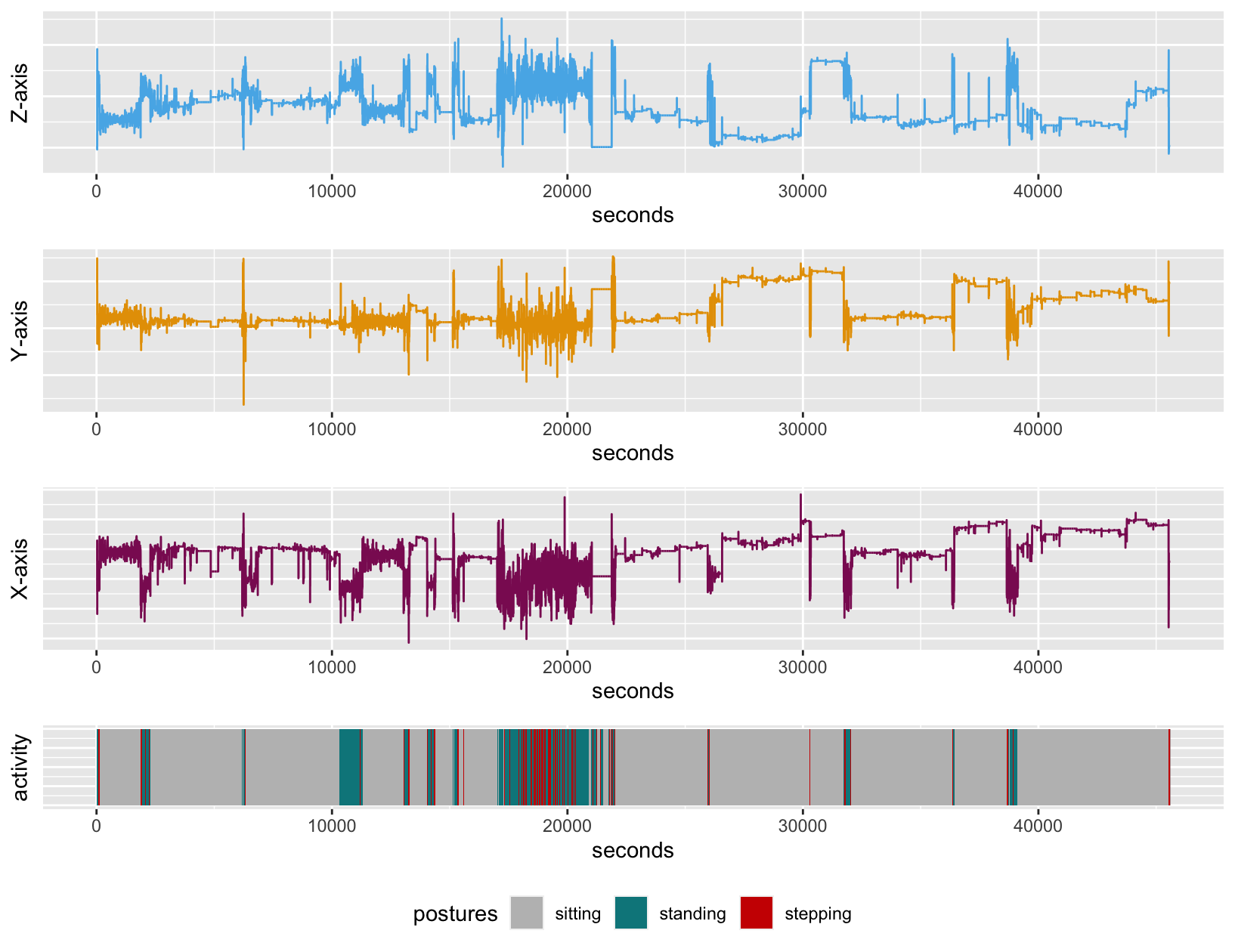}
		\caption{An example of a daily file plot.}
		\label{figure:1}
	\end{figure}

	
	We remove records identified as non-wear times, mostly due to sleeping, from the 7-day  ActiGraph GT3X+ and activPAL data using the Choi algorithm \cite{CHOI2012}.
	After removing the non-wear times, the datasets are separated into roughly
	seven pieces, where each segment typically represents a long continuous wearing epoch for each day. 
	These segments, which we call "daily files", are the fundamental blocks for later data analysis.
	In addition, the gravity component of the ActiGraph data (details in Section 3.1) is preserved in the processing as it serves a key role for our estimation method.
	As an example, in Figure \ref{figure:1} we provide an example of daily file of one participant. 
	
	
	We further break the raw data into 
	a collection of (data, label) tuples at the {\it changepoints} of posture labels.  Here, a {\it changepoint} is the time stamp at which a posture change occurs. 
	Therefore, by doing this we will have a collection of training samples where each sample contains data of a \textit{bout} (period between two consecutive change-points) with consistent posture. 
	The bout length information is also retained for subsequent analysis.

	\section{The Model}
	
	We describe in detail the proposed local Markov switching model.
	
	\subsection{Model setup}\label{sec: 3.1}

	The accelerometer of the Actigraph GT3X+ measures the current total acceleration of the device, relative to its local free-fall inertial frame. Due to gravity, the accelerometer at rest records an acceleration of 1g (standard gravity), pointing to the surface of the earth. 
	As for both sitting and standing bouts, the device stays "at rest" most of the time, and we cannot distinguish between these two states using vector magnitude (VM $=\sqrt{x^2 + y^2 + z^2}$) alone.

\begin{figure}[h]
		\centering
			\includegraphics[scale = 0.4]{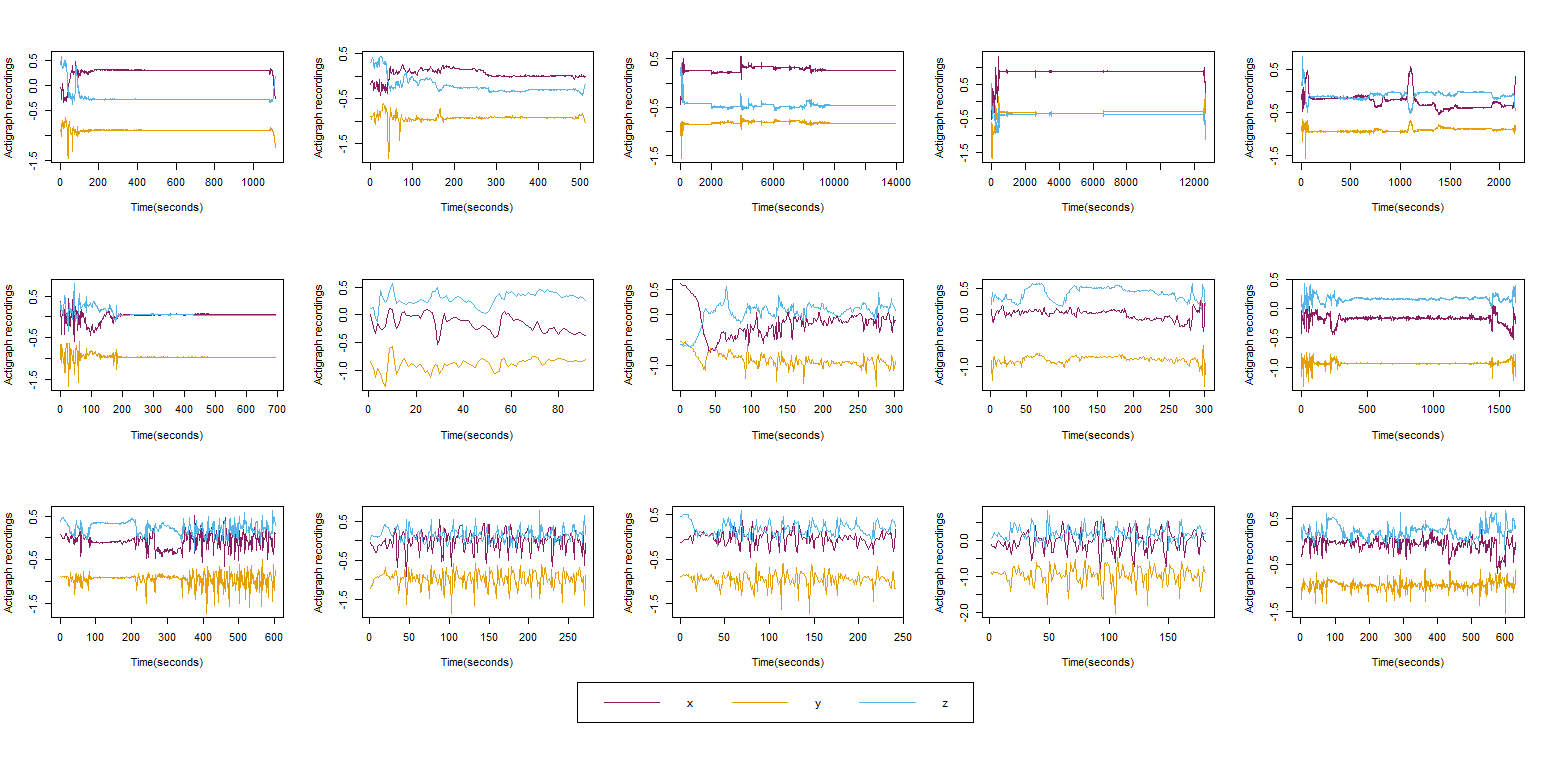}
		\caption{Actigraph plots for different postures in one daily file. Top row: sitting; middle row: standing; bottom row: stepping. }
	\label{fig:bouts}
	\end{figure}

    Instead of relying on the vector magnitude alone, we model the triaxial time series of the ActiGraph records, which contain information on both the magnitudes and the {\it directions} of device acceleration.
	The triaxial time series record shifting of the acceleration direction, which is a major characteristic factor for transitions between sitting and standing,
	thus are capable of separating of sitting and standing postures. In addition, for continuous stepping bouts, movement of the device mostly fluctuates periodically around its standing position, and stepping activities induce a random noise component with larger variance than resting situations. An example daily file illustrating the above observations is given in Figure \ref{fig:bouts}. 
	
	A free-living situation can result in the hip-worn device's position varying between participants and wearing epochs. Accounting for this variation, we only assume that within each wearing epoch, the position of the device does not change over time. 
	Upon examining the data, we found that most of the daily files abide by this rule, while there are a few data points that violate it, as a result of the device being moved around the hip or flipped to face a different direction.

	Based on the observations above, we model the Actigraph GT3X+ accelerometer data with a simple Markov switching model where the the underlying posture follows a (hidden) Markov process. 
	Formally, let $\mathbf{X}_{i,j}(t)$ denote the tri-axial Actigraph data at time point $t$ of participant $i$ at wearing epoch $j$ for $j = \{1,2,\cdots,7\}$. 
	$\mathbf{X}_{i,j}(t)$ is a triaxial time series for each $i$ and $j$.
	Let $L_{i,j}(t)$ denote the hidden posture label at time $t$ for participant $i$ at wearing period $j$.
	
	We adapt the classical decomposition of a time series:
	\begin{equation}
	\mathbf{X}(t) = \boldsymbol{\mu}(t) + \mathbf{S}(t) + \boldsymbol{\epsilon}(t),
	\end{equation}\label{eq1}
	where $\boldsymbol{\mu}(t)$ is the trend component, $\mathbf{S}(t)$ is the seasonal component, and $\boldsymbol{\epsilon}(t)$ is a stationary time series with mean $\mathbf{0}$. Both $\boldsymbol{\mu}(t)$ and $\mathbf{S}(t)$ are deterministic in this decomposition.
	
	Incorporating the times series model with a Markov switching model, we present our model for the triaxial Actigraph data as follows:
	
	\begin{equation}
	\label{eq2}
	\mathbf{X}_{i,j} (t) =
	\begin{cases}
	\boldsymbol{\mu}_{i,j}^{(0)} + \boldsymbol{\epsilon}^{(0)}_{i,j}(t) & \text{if $L_{i,j}(t) = 0$},\\
	\\
	\boldsymbol{\mu}_{i,j}^{(1)} + \boldsymbol{\epsilon}^{(1)}_{i,j}(t)  & \text{if $L_{i,j}(t) = 1$},\\
	\\
	\boldsymbol{\mu}_{i,j}^{(1)} + \mathbf{S}(t) + \boldsymbol{\epsilon}^{(2)}_{i,j}(t)  & \text{if $L_{i,j}(t) = 2$},
	\end{cases}       
	\end{equation} \label{eq: model}
	with activity codes $L_{i,j}(t) = 0,1,2$ representing sitting, standing and stepping postures correspondingly. 
	
	Model \eqref{eq: model} takes into account several key aspects of the accelerometer data that are realistic in real world scenarios, as  elaborated below:
	\begin{enumerate}
	
    \item The model is "local" in the sense that it is parametrized differently for each wearing epoch $(i,j)$.
    Consequently, model parameters for unseen test data cannot be fully inferred by exploiting the characteristics of training data;
	also, the classical Baum-Welch algorithm \cite{baumwelch} for parameter fitting, and the Viterbi algorithm for sequential hidden states prediction fail to meet the assumptions of our model, and will not work in general.

	\item The components $\boldsymbol{\mu}^{(0)}_{i,j}$ and $\boldsymbol{\mu}^{(1)}_{i,j}$ are constant trend components representing the opposite direction of gravity. 
	For different participant $i$ and wearing epoch $j$, $\boldsymbol{\mu}_{i,j}$ are allowed to be different, as the position of device can vary between participants and wearing epoch.
	
	\item  The model reflects the observation from raw data plots (see Figure \ref{figure:1}) that the major difference between sitting and standing bouts is the mean component $\boldsymbol{\mu}$, and a {\it mean-shift} in the data occurs when a sit-stand transition occurs. 
	
	\item The standing and stepping bouts are modeled to share the same constant trend $\boldsymbol{\mu}_{i,j}^{(1)}$ as movements of the device during stepping fluctuate around its standing position.	
	For stepping bouts, however, an additional non-random seasonal component $\mathbf S(t)$ is brought into the picture to characterize the fluctuation caused by the stepping movements. 

	\item Each of $\boldsymbol \epsilon^{(0)}(t)$,$\boldsymbol \epsilon^{(1)}(t)$ and $\boldsymbol\epsilon^{(2)}(t)$, which are errors not explained by the trends nor the seasonal component, is assumed to be uncorrelated across time $t$, or independent and identically distributed (iid), and is assumed independent across different $(i,j)$.
	Note that $\boldsymbol \epsilon^{(0)}(t)$, $\boldsymbol \epsilon^{(1)}(t)$ and $\boldsymbol\epsilon^{(2)}(t)$ are not assumed to have the same distribution. In fact, it is observed from the data that $\boldsymbol\epsilon^{(2)}(t)$ has variance significantly larger than $\boldsymbol \epsilon^{(0)}(t)$ and $\boldsymbol \epsilon^{(1)}(t)$ across all $i,j$, while $\boldsymbol\epsilon^{(1)}(t)$ only has slightly higher variation than $\boldsymbol\epsilon^{(0)}(t)$. See Figure \ref{fig:bouts} for an illustration.
	\end{enumerate}

	\subsection{Local and Global Data Features}
	For the purpose of model fitting, meaningful features  are those that are  representative of the whole data, including both training and test sets. 
	Features extracted from the accelerometer data can be categorized into the following types:
	A {\it local feature} is defined as extractable feature of data from a  time interval during which its posture label is constant.
	A {\it global feature} is defined as useful extractable information concerning the whole range of data $\mathbf X_{i,j}(t), t = \{1,\cdots,T_{i,j}\}$, where $T_{i,j}$ is the number of total observations for daily file $(i,j)$. For example, the mean and variance of each pre-segmented bout are local features. 
	The mean-shift phenomena however, is not observable from looking at an interval with a constant posture label, therefore it is a global feature. 
	The range of probable bout lengths of different postures is another global feature that can only be observed by looking at the entire sequence.
	
	The form of Model \eqref{eq: model} suggests that we may train an accurate classifier using local features for distinguishing stepping from non-stepping postures, including both sitting and standing, through proper feature engineering that captures seasonal and variance information. In fact, the random forest classifier recorded a 90\% accuracy for identifying stepping and non-stepping bouts. 
	
	However, the seasonal and variance features are insufficient for further separation of non-stepping posture into sitting and standing. Model (\ref{eq: model}) suggests that 
	the main difference that separates
	the two postures are their mean components, which are parametrized differently for each daily file $(i,j)$. Because of this local parametrization, it is possible that a sitting mean component in the training set is also a standing mean in the test set, and using the mean information as a feature will not lead to good classification results. 
	

	We show in the following section that by incorporating global features in a classification algorithm,  further separation of sitting/standing bouts can be carried out with high accuracy.
	Alas, global features of a test set are not directly calculable without some label information. This means we need to be able to predict certain label information for a test set before calculating global features. These insights informed our novel approach, described  below; we note also that  our method achieves good classification accuracy.

\begin{figure}[h]
		\centering
		\includegraphics[scale = 0.32]{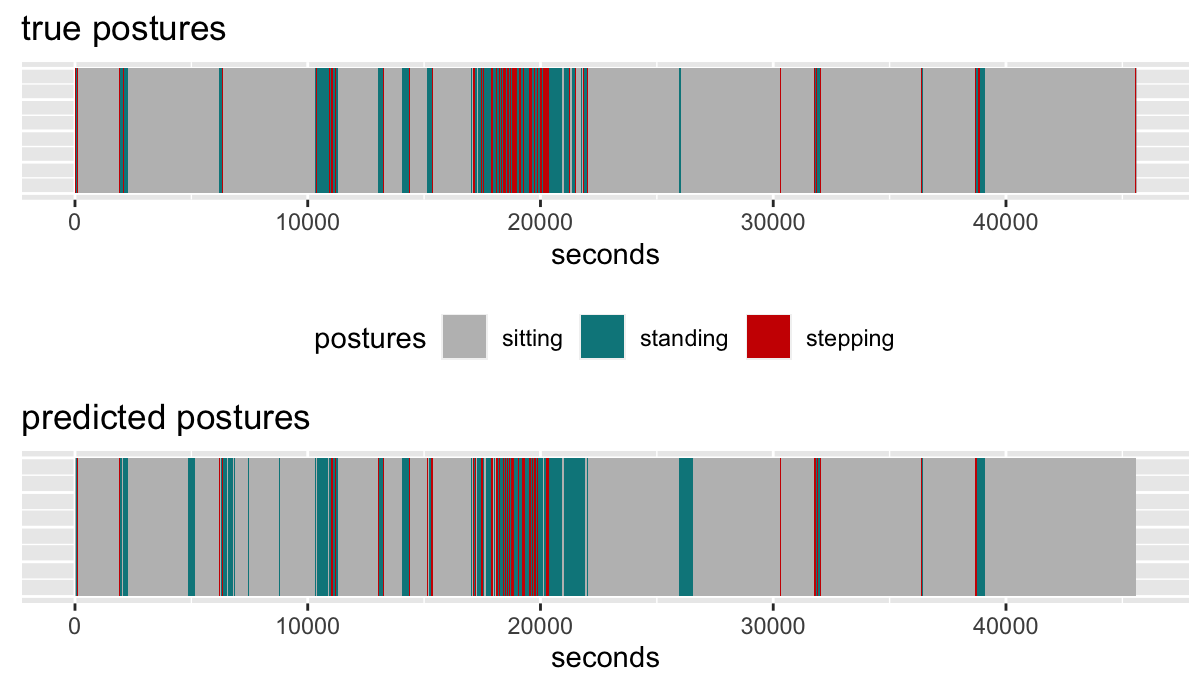}
		\caption{Prediction result of figure 1.}
		\label{figure:2}
	\end{figure}

	In the prelude of section 1, we briefly described our approach in three steps. Here we provide a detailed summary. For each daily file $(i,j)$:
	\begin{enumerate}
		\item The whole observation is split into contiguous segments
		such that each segment is believed to have a fixed posture label. Under equation (2), this task is achieved with 
		{\it changepoint detection} techniques, and it offers several advantages compared to previous approaches which use fixed-length bouts. 
		The classification task will be performed on each segment obtained with the change-point detection.
		\item We then perform classifications with a two stage procedure. In the first stage, a classifier trained with local features from the training set splits the segments into
		stepping and non-stepping bouts. Then, with the additional stepping bout information, we use the mean-shift global feature to assist further separation of
		sitting and standing bouts under a multiple testing setup. 
		\item Finally we use the range of sitting/standing bout lengths to make post correction for spurious classification. 
	\end{enumerate}
	As a demonstration, the prediction result of data in Figure \ref{figure:1} using our approach can be found in Figure \ref{figure:2}. 
	In the following subsections, the details and implementation  of  each step are provided. 

\section{Multistage classification method}
    In what follows, we describe the multi-stage method for segmenting and classifying sedentary bouts (i.e., sitting, standing, steppin) using the triaxial data collected by the  ActiGraph GT3X+ monitor.
    
	\subsection{Changepoint Detection}\label{sec4.1}
	Detecting sudden changes in the data generating process falls into the realm of changepoint detection for time series. 
	This can be achieved training-free with the help of statistical analysis; or training-based using machine learning methods. 
	A survey can be found in \cite{pmid28603327}. 
	
	For our purpose, we need a method that can detect the "structural" changes for the model in equation (2), and more importantly 
	executes quickly for large scale data.  
	The structural changes can be summarized into the following: mean-shift for sit-stand transitions and variance change for stand-step transitions,
	and then a combination of both for sit-step transitions.
	There are existing methods for detection of mean shift and variance change. Under a multivariate setup, for example, the {\it ecp} package  in \textsc{R} is capable of 
	detecting multiple changepoints with any changes in marginal distribution \cite{JSSv062i07}. In the univariate case, the standard \textsc{R} library
	{\it changepoint} of \cite{JSSv058i03} offers detection of changes in mean and variance; it also offers a wide range of optimization algorithms
	suitable for multiple changepoint detection with different penalty selection. 
	
	Simulation results show both {\it ecp} and {\it changepoint} packages can detect changepoints with high sensitivity, while also having a noticeable high false discovery rate (FDR), thus shattering the true segments into smaller pieces. A high sensitivity is crucial to guarantee the true changepoints are detected, so that each segment will only contain data from a single posture; while false discoveries can be tolerated if the classification algorithm can still perform well on smaller segmennts.
	
	However, even with a high level parallelization, {\it ecp} takes a long time (i.e., upto 2 hours for a single daily accelerometer file) to produce results for a single daily file which makes it less attractive for large scale
	implementation, without access to high-performance computer clusters. As a result, we choose the {\it changepoint} package and use it on the axis which is the direction  that exhibits the most substantial behavior change, as is evident from 
	visual inspection. Typically, the $x$-axis is the preferred choice. 
	
	An important detail worth mentioning is pre-segmentation and parallelization.
	As the time complexity of a single changepoint detection is $\mathcal O(T^2)$, $T$ being the length of a sequence, the time cost scales quadratically as sequence length increases, and computing changepoints for each daily file becomes intractable due to the sheer size of our data. 
	To work around this problem,  we manually segregate the data into fixed length segments and detect changepoints in parallel within each segments. The results are concatenated
	to form a complete sequence of changepoints for the whole data. This will introduces some false discoveries, but will dramatically reduce computational cost. 
	The length of segments is determined by cross validation to be introduced in the next section. The detected changepoints are then used to split the entire sequence as consecutive segments with undetermined hidden labels.

	\subsection{Two stage classification}
	We adapt the random forest classifier used in \cite{pmid29443824} to our varying length scenario such that
	a first stage separation of stepping bouts from sitting and standing bouts is performed with high accuracy. The detected stepping bouts can then be used to perform inference on parameters in model \ref{eq: model}. Then, a second stage classification can be performed to further separate non-stepping bouts with this additional information.
	
	\subsubsection{Detecting stepping bouts with random forest model}
	Details of training the classifier are as follows. The training set comprises of data that are segmented by the actual changepoints of hidden labels. The following features are extracted from each training set to form a feature vector with fixed length that are inputs
	to the fitting procedure: The mean, SD, coefficients of variation, minimum, maximum, and 25th, 50th, 75th percent quantiles of the VM sequence; 
	position of maximum absolute autocovariances of VM up to lag 25; correlations between each axis; the roll, yaw, pitch angles of the direction of acceleration
	with $\text{roll} = \text{tan}^{-1}(y,z)$, $\text{yaw} = \text{tan}^{-1}(y,x)$ and $\text{pitch} = \text{tan}^{-1}(x,z)$; The direction of gravity component is computed as the roll, yaw, pitch angles
	of the mean values of 3 axis after applying a low-pass filter of 0.5HZ. Periodograms for the VM series are calculated with the following features extracted 
	that describe hidden periodicities: frequency where maximum of periodogram occurred; the value of periodogram at this frequency; frequency between 0.3 HZ and 
	3 HZ where maximum of periodogram in this band occurred and the value of periodogram at this frequency. Finally, the entropy of the frequency domain signal is
	calculated as a measure of uncertainty. 
	We prepare the training samples by performing the above-mentioned feature extraction on the data tuples mentioned in section 2.2. 
	
	The raw features along with labels representing stepping and non-stepping are used to train a random forest model with 500 decision trees. 
	While the random forest model is trained with segmented series with true changepoints, its generalization to segments split by the estimated changepoints
	is more important to our analysis. Therefore, the test errors of the random forest model with  segmentation performed under both true changepoints and estimated changepoints are used as the standards in  evaluating the performance of the classifier.

We perform training and testing of the random forest model on disjoint sets of participants detailed in the following: 
15 participants are randomly split into training and testing pools by a 2:1 ratio, and the training/testing samples are sampled from each pool.
We apply balanced sampling to ensure the sample sizes of each posture are approximately equal in the training set. The number of (features, label) pairs in the training and testing sets are $15000$ and $2000$ respectively.
Our results indicate that the trained classifier has 10.7\% non-step to step error rate (fraction of non-stepping bouts classified as stepping)  and 7.3\% step to non-step error rate for 
a test set segmented by true changepoints; while surprisingly, for test set segmented by detected changepoints, the non-step to step error is only 2.0\%, with
step to non-step error increased to an average of 15.4\%.

	Certain parameters of the changepoint detection algorithm affect the classifier's performance on the changepoint-produced segments. Most notable are the window size for parallel changepoint detection (windowwidth)
	and minimum length between two changepoints (minseglen). We use leave-one-out cross validation (LOOCV) to select these two parameters that minimize the following quantity:
	for each daily file, define
	\begin{equation}
	\textit{Sit-to-step error} = \frac{\textit{number of identified stepping bouts with true label sitting}}{\textit{number of identified stepping bouts}}.
	\end{equation}
	The sit-to-step error represents the percentage of sitting bouts among the identified stepping bouts, and minimization of this quantity is crucial for the success of second stage classification that separates sitting and standing bouts. In second stage classification, the distribution of mean information from predicted stepping bouts are used to calculate the 'confidence interval' for the mean of standing bouts, using the fact that standing and stepping bouts share the same mean in equation (\ref{eq: model}). Some portion of the stepping mean information is inevitably mixed with standing means and sitting means as the first step classification procedure is not 100\% accurate. The mixture of standing means does no harm as the standing/stepping mean are assumed to be the same; while the sitting mean will contaminate the mean information and potentially sabotage the second stage classification. In other words, in the language of robust statistics, this error represents the contamination fraction in the standing/stepping mean distribution by the sitting mean distribution, and
	we require minimum mixture of sitting bouts in first stage classification to control the contamination level.
	Using LOOCV with a grid search on parameter values shows that $windowwidth = 1800$ and $minseglen = 450$ achieves minimal sit-to-step error.
	
	\subsubsection{Local multiple testing for second stage classification}
	The mean shift between sitting and standing bouts as a global feature is our main focus to perform further separation between sitting and standing bouts.
	With the help of identification of stepping bouts from first stage classification, a hypothesis testing approach for mean equality is designed for further separation. Let $\boldsymbol{\mu}_{i,j}^{(b)}$ be the mean vector of a non-stepping bout $b$ from daily file $(i,j)$, our hypothesis test is the following:
	
	$$H_0: \boldsymbol{\mu}_{i,j}^{(b)} = \boldsymbol{\mu}_{i,j}^{(1)}$$
	
	vs.
	
	$$H_1: \boldsymbol{\mu}_{i,j}^{(b)} \neq \boldsymbol{\mu}_{i,j}^{(1)}$$
	and we assign the non-stepping bout $b$ as sitting when the test is rejected.
	
	Note that this test is specifically proposed for testing within the $(i,j)$th daily file and thus is local.
	As a result, the rejection region of the test also needs to be designed specifically for $(i,j)$. 
	We apply the following empirical rule, which is proven to be effective in practice. 
	Let $\widehat{\mathbf F}_{i,j}$ be the cumulative distribution function(CDF) of the mean vectors
	from detected stepping bouts within the $(i,j)$th daily file.
	Since we assume equality of the mean components for standing and stepping bouts,
	the $(1-\alpha)$ confidence region generated from  $\widehat{\mathbf F}_{i,j}$ should contain $\boldsymbol{\mu}_{i,j}^{(1)}$ with a high probability.
	On the contrary, the sitting mean $\boldsymbol{\mu}_{i,j}^{(0)}$ is unlikely to be in this region, since sitting posture is outside the range of
	movement for a stepping bout. Here, $\alpha$ is a universal parameter that has a pivotal role in controlling accuracy of this test which should also be determined by cross validation. Our results show that $\alpha = 5\%$ actually has good performance for classification accuracy and total sedentary time estimation.
	For implementation purpose, the above idea is simplified to construct confidence interval using the mean information on the axis with the most significant mean-shift behavior. The axis of choice is the same 
	as the one chosen in changepoint detection. 
	
	\subsection{Spurious label correction by thresholding} 
	After the second stage of classification, the prediction results are agglomerated 
	and adjacent bouts with same predicted label are concatenated to form a new stream of bouts with distinct adjacent labels, a representation of the predicted posture/activity log from data that is analogous to the thigh-worn ground truth.
	The length of non-stepping bouts are calculated and compared with the range of lengths of sitting/standing bouts of the training data for post-correction. 
	In particular, we found that only 3\% of the standing bouts from training set have length larger than 3 minutes, while none have length larger than 10 minutes. 
	Therefore, we make a reasonable correction to the results by the following rule:
	a standing bout from predictions will be corrected to sitting if its lengths is larger than some stand threshold. The threshold is found by cross validation with the aim of minimizing classification error. For this dataset, a 10-min stand threshold produces optimal results.
	

	\section{Experimental Results}
	A primary gap in sedentary behavior measurement research is the lack of posture classification, specifically sitting behavior. Hence with a primary goal of gauging the accuracy of our method for classifying sitting postures,  our results are summarized in three different aspects. First of all, we compute the confusion matrix for sitting versus non-sitting labels for the entire population, measured in number of seconds.
	Second, we compare the predicted total sitting time to the ground truth for each participant to measure the accuracy of our method at the individual person level. 
	Finally, we demonstrate how our prediction results can be used to accurately estimate statistics related to sedentary behavior  through closeness of the predicted distribution of sitting bout lengths versus the ground truth. We note that in these comparisons, the predictions of our algorithm are based on test data that were not used in training the algorithm.

	\begin{table}[h]\label{table:1}
		\centering
		\caption {Confusion matrix of classification results(in seconds)}
		\pgfplotstabletypeset[
		every head row/.style={%
			before row={\toprule 
				& \multicolumn{2}{c}{ActivPAL reference}\\            \cmidrule{2-3}},
			after row=\midrule},
		every last row/.style={after row=\bottomrule},
		columns/Prediction/.style={string type},
		columns/Sitting/.style={string type},
		columns/Non-Sitting/.style={string type},
		columns/Prediction total/.style={string type},
		]{
			Prediction       Sitting           Non-Sitting    {Prediction total}
			Sitting          $44.6\times 10^5$    $9.2\times 10^5$  $53.8\times 10^5$
			Non-Sitting      $8.3\times 10^5$    $28.0\times 10^5$  $36.3\times 10^5$
			{ActivPAL total}  $52.9\times 10^5$   $37.2\times 10^5$  $90.1\times 10^5$
		}
		
	\end{table}
	
	Table 1 contains the classification results compared with the activPAL ground truth of all participants measured in number of seconds. 
	The following metrics are also helpful for measurement purpose: sensitivity, 84.3\%; specificity, 75.2\%; balanced accuracy: 79.7\%;
	precision, 82.9\%; F1-score, 83.6\%.
	In \cite{KERR2013290}, the sensitivity and specificity for sitting posture are 64.2\% and 69.2\% respectively, showing that our method performs favorably. 
	
\begin{figure}[h]
		\centering
		\includegraphics[scale = 0.3]{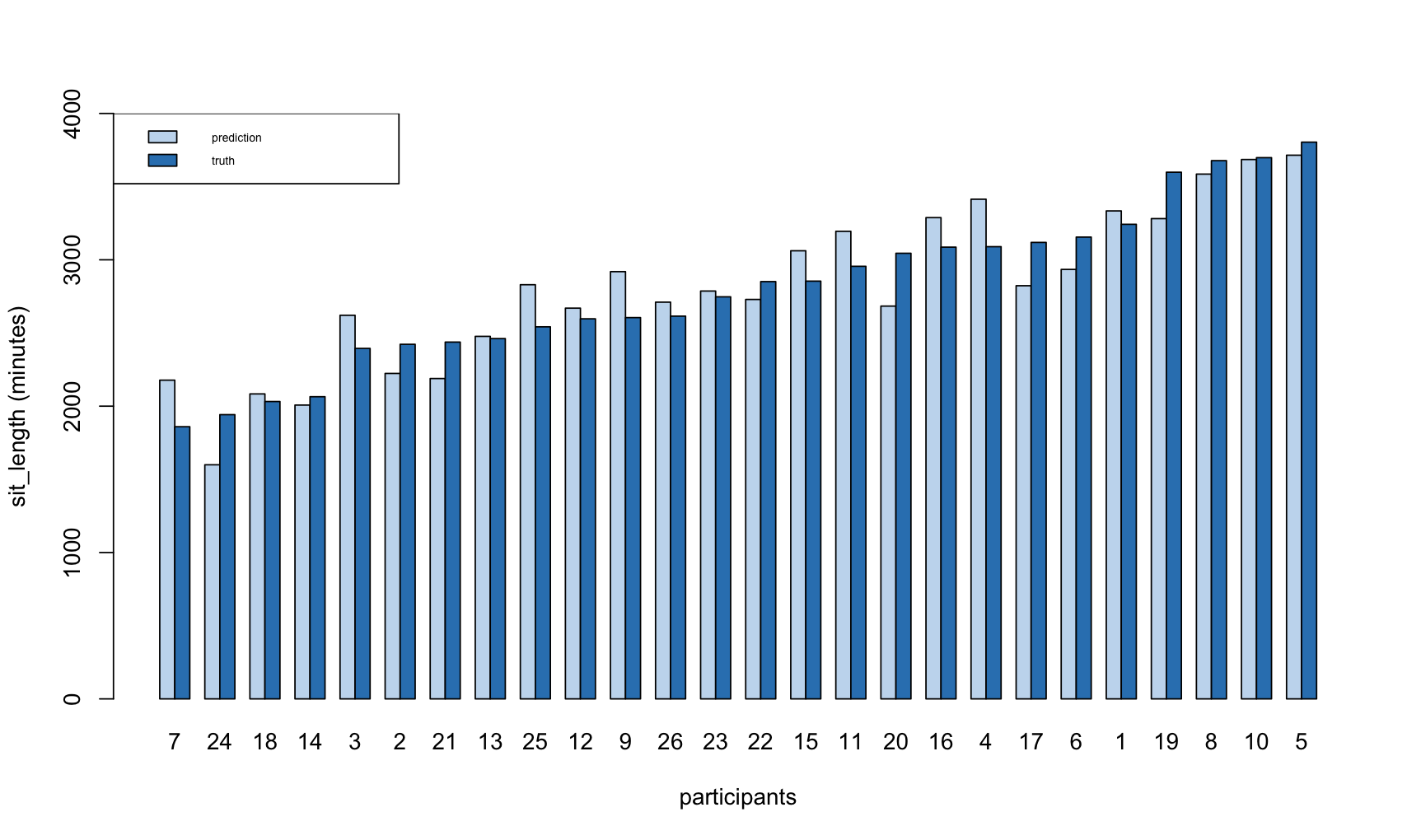}
		\caption{Total sedentary time estimates(in minutes) for 26 participants.}
		\label{figure:3}
	\end{figure}

\begin{figure}[h]
		\centering
		\includegraphics[scale = 0.39]{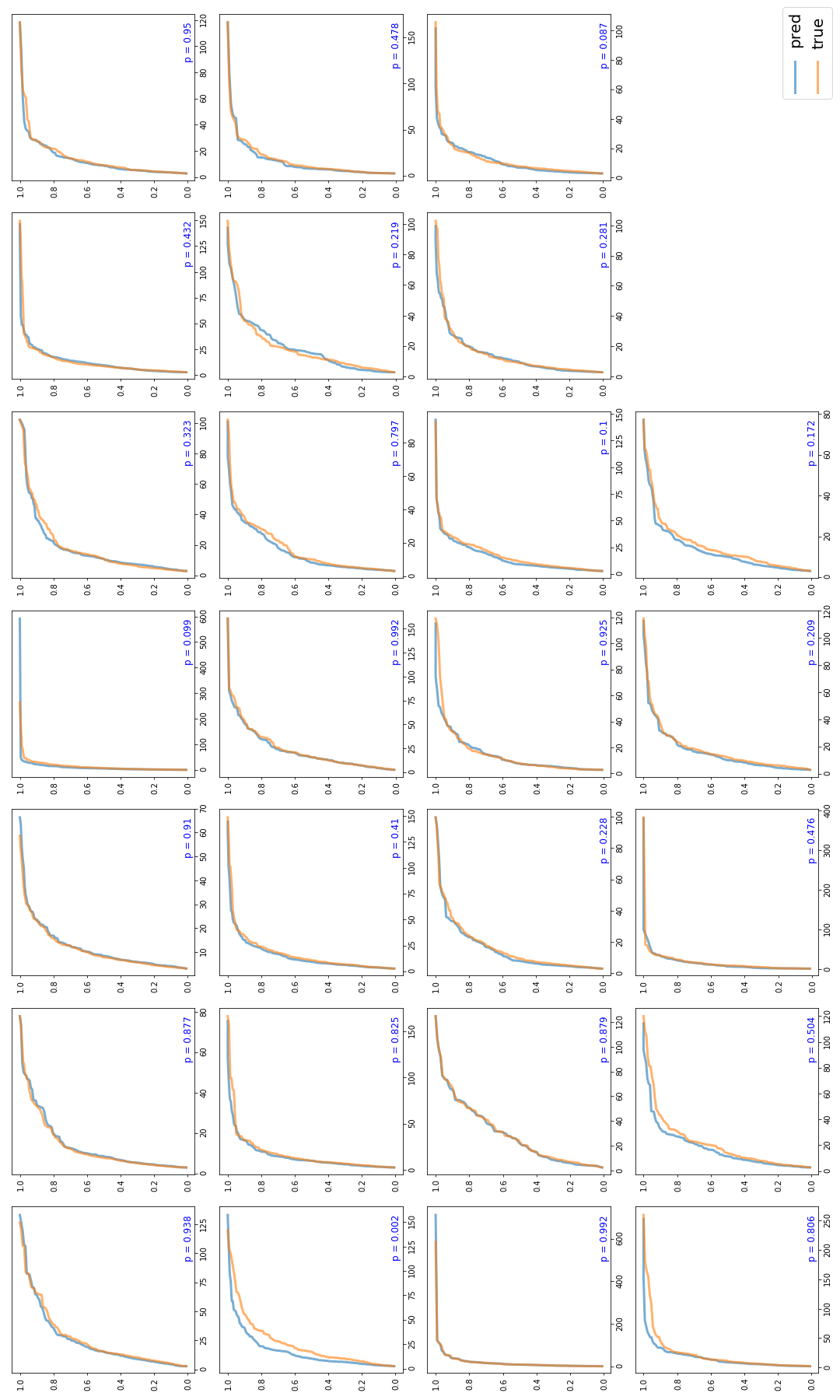}
		\caption{Comparison of ECDF between predicted length of sit bouts vs. truth. }
		\label{figure:4}
	\end{figure}

	Figure \ref{figure:3} contains measures of accuracy for total sitting time estimates of 26 participants. 
	The plot shows that our algorithm can track the actual sedentary times with good accuracy for all participants. Therefore it is well suited for estimating person-level total sitting time. Finally, the algorithm can also estimate prolonged sitting patterns. This is illustrated in the figure \ref{figure:4}, where we plot the cumulative distribution functions (CDF) of estimated sitting bout lengths (in minutes) along with the ActivPAL references, and also provide the $p$-values from the Kolmogorov-Smirnov goodness-of-fit test. 
	We consider sitting bouts with length longer than a 3-minute threshold. 
	These CDFs can be used to calculate 
	a wide range of characteristic statistics that measure prolonged sitting patterns, see \cite{pmid19854651} for examples of such statistics. By demonstrating high concordance between the CDFs, we can be confident that derived statistics of sitting patterns calculated using our predictions will be close to their true values as well. 
	
\section{Discussion}
In this paper, we proposed a new approach for the classification of hip-worn accelerometer data to sit/non-sit postures under free-living situations. The approach produces accurate classification results and also meaningful sedentary behavior estimation compared to previous solutions. In this section, we summarize the main components, limitations and future directions of our work. 
\begin{enumerate}
    \item \textbf{Our multistage classification method and potential scope for improvement.} Our approach offers a general framework consisting of multiple algorithms in different stages: changepoint detection; the random forest algorithm in first stage classification; mean-shift hypothesis testing in second stage classification; post-correction algorithm.
    Moreover, the framework is very flexible as we can adopt more advanced algorithm choices at different stages of classification than what is used here. 
    We mention a few scenarios where further improvements can be made for each of the algorithms above. 
    \begin{itemize}
        \item {\it High FDR for changepoint detection.} As noted in section \ref{sec4.1}, results from both {\it ecp} and {\it changepoint} algorithms suffer from a high FDR, making the segmented series not identical to those produced by true changepoints, and therefore potentially affecting the performance of classification in later steps. The discrepancies arise partially because the idealistic assumptions of these methods are inconsistent with the noisy nature of the actual data. On the other hand, machine-learning based algorithms \cite{pmid28603327} can be trained for detecting changepoints as well, and have the potential of being more robust against model misspecification, making them a better candidate for changepoint detection. Even so, our simple model-based assumptions which are visually verifiable, make our approach easier to interpret and plausible. 
        
        \item {\it Deep learning based algorithm for first stage classification.}  Neural networks such as CNN can automatically select meaningful features from raw input data without feature engineering, which could potentially boost classification performance. 
        In the recent work of \cite{Nakandala2020}, a CNN architecture is 
        proposed to classify 3-second window observations into sitting, standing and stepping. 
        They showed that CNN achieves higher accuracy than classical machine learning methods, such as logistic regression and random forest methods. 
        Their best result in terms of balanced accuracy is 83.3\%, which is slightly higher than ours.
        However, their result is not directly comparable with ours, as the granularity level in \cite{Nakandala2020} is 3 seconds where the data considered must have consistent label in each window; on the other hand, our accuracy is calculated at the finest level, 0.1 second.

        \item {\it Mean-shift hypothesis testing using triaxial data.} Our implementation in the second classification stage involves producing confidence intervals for the standing mean in one axis -- one with the most significant "mean-shifting" phenomena. Even with this selection, we have found that there are some data whose degree of "mean-shifting" is not significant in any axis. In this scenario, it is natural to consider using information of all three axes. However, the choice of "confidence region" for the mean {\it vector} is more complicated as there isn't a natural ordering in three dimensions. As a result, different shapes of confidence regions need to be considered, and we leave this to future work with larger number of samples. 
    \end{itemize}
    
    \item \textbf{Validity of the proposed model}. 
    The proposed model \ref{eq: model} is based on our observation of the data and also the theoretical behavior of hip-worn accelerometer under different postures. However, we also observed a few data points which exhibit discrepancies between the proposed model and actual data. 
    Examples for the violation of model assumptions include (1) gradual  shifting and rapid rotation of hip-worn device causing shifting of mean component within one daily file; (2) inseparable sit/stand mean components due to poor device positioning;
    (3) significant noise during sitting epochs due to various reasons; etc.  While we were successful in addressing  these issues to some degree in the post-correction step, there could be more dedicated algorithms for resolving the above irregularities. For example, for (1) and (2) we have experimented with an algorithm utilizing a $3\times 3$ rotation matrix such that the rotated data have both stationary and maximum separability of different mean components. Again, we leave further consideration of these geometric transformations to future work with larger number of samples.
    
    \item \textbf{Generalization to other datasets}. The dataset we use comprises of  older women who are breast cancer survivors. We plan to investigate if our approach can be generalized for a different population, such as different age, biological sex or race and ethnic groups. 
    
\end{enumerate}

\section{Conclusions}
In this paper, we investigated the problem of posture prediction for hip-worn accelerometer data under free-living situations. We proposed a general framework to tackle this task by means of segmentation with changepoint detection and two-stage classification using machine learning and hypothesis testing. Our method achieves good prediction results in terms of prediction accuracy and accuracy of sitting behavior estimation; and it is advantageous in terms of interpretability. Further investigation includes alternative algorithm choices at different steps of our methods and generalization to more complex datasets.

\bibliography{references}

\end{document}